\documentclass[runningheads]{llncs}

\def\*#1{\mathbf{#1}}
\def\+#1{\mathcal{#1}}
\def\-#1{\mathbb{#1}}
\def\~#1{\mathrm{#1}}
\def\R{\mathbb{R}}

\renewcommand{\hbar}{\bar{h}}

\usepackage[english]{babel}
\usepackage{amsmath, amssymb, bm}
\usepackage{algorithm, algcompatible}
\usepackage{rotating}
\usepackage{multirow}

\usepackage{booktabs}

\usepackage[T1]{fontenc}
\usepackage{caption}    
\usepackage{graphicx,verbatim}
\usepackage[table]{xcolor}
\usepackage[colorlinks=true,
            linkcolor=blue,
            citecolor=blue,
            urlcolor=blue]{hyperref}

\usepackage{orcidlink}

\usepackage{pgfplots}
\usepackage{pgf}
\usepackage{tikz}
\usetikzlibrary{matrix}
\usetikzlibrary{positioning, arrows.meta, calc, spy}

\usetikzlibrary{fit}					  
\usetikzlibrary{backgrounds}	          
\usetikzlibrary{shapes.arrows}
\usetikzlibrary{shapes.geometric}
\usetikzlibrary{shapes.multipart}
\usetikzlibrary{matrix, arrows, positioning}
\tikzstyle{rec}=[draw,rectangle, minimum height=2cm]
\tikzset{>=stealth', punkt/.style={
		fill=gray!40, 
		draw=black, very thick, text width=4.7em, minimum height=3em, text centered}}
\tikzset{>=stealth', Denoi/.style={rectangle, fill=blue!20, 
		draw=black, very thick, text width=6em, minimum height=3.5em, text centered}}
\tikzset{>=stealth', CG/.style={rectangle, 
		fill=green!20, draw=black, very thick, text width=9.5em, minimum height=3.5em, text centered}}

\tikzstyle{background} = [rectangle, fill=green!20, inner sep=0.1cm, rounded corners=4mm, 
minimum height=6em]
\tikzstyle{sum}   = [draw, fill=gray!40, circle, node distance=1cm]
\tikzstyle{dot}   = [circle, fill=black, inner sep=0pt, minimum size=5pt, node contents={}]
\tikzstyle{fig_n} = [node distance=30pt, inner sep=0cm]

\begin{document}

\title{EPC-3D-Diff: Equivariant Physics Consistent Conditional 3D Latent Diffusion for CBCT to CT Synthesis}
\titlerunning{EPC-3D-Diff: Equivariant Physics Consistent Diffusion}

\author{Alzahra Altalib\inst{1,2}\orcidlink{0009-0009-6082-9754}, Chunhui Li\inst{1}\orcidlink{0000-0003-2186-5137}, Haytham Ahmad Alewaidat\inst{2}\orcidlink{0000-0001-6046-9570}, Khaled Z. Alawneh\inst{2}\orcidlink{0000-0002-2615-1988}, Ahmad Awad Qandeel\inst{3}, Alessandro Perelli\inst{4}\orcidlink{0000-0002-0511-2293}}  
\authorrunning{Alzahra Altalib et al.}
\institute{School of Science and Engineering, University of Dundee UK \email{2600129@dundee.ac.uk} \and Faculty of Applied Medical Sciences, Jordan University of Science and Technology \and Experia Healthcare, Jordan \and School of Cardiovascular and Metabolic Health, University of Glasgow UK\\
    \email{alessandro.perelli@glasgow.ac.uk}}

\maketitle              

\begin{abstract}
Cone-beam CT (CBCT) is routinely acquired during radiotherapy for patient setup, but its quantitative reliability is degraded by scatter, noise, and reconstruction artifacts, limiting Hounsfield Unit (HU) accuracy. 
We propose EPC-3D-Diff, a novel conditional 3D latent diffusion framework for volumetric CBCT to CT synthesis that introduces a projection domain equivariance loss derived from acquisition physics. Unlike common image domain equivariance, we exploit the fact that an in plane rotation of the volume corresponds to an angular shift in its projections. During training, we enforce this relationship by forward projecting rotated synthesized CT volumes and matching them to appropriately angle shifted projections of the paired target CT, yielding a physics consistent equivariance constraint integrated into the diffusion objective.
To capture full 3D context efficiently, conditional diffusion is performed in a compact latent space learnt by a lightweight 3D autoencoder, preserving axial depth while downsampling in plane resolution for stable training. 
We validate on a paired head CBCT/CT phantom dataset, including repeat scans, and paired clinical data using patient wise splits, and perform single and mixed domain training, ablations, and comparisons with diffusion and CycleGAN. EPC-3D-Diff generalizes well and achieved substantial improvements, $+7.4$ dB (phantom) and $+1.8$ dB (clinical data) in PSNR compared to state of the art methods, alongside improved SSIM and HU accuracy, within tissue boundaries. Overall, EPC-3D-Diff improves robustness and physics consistency, supporting HU aware synthesis for downstream radiotherapy workflows. The open source code for EPC-3D-Diff is available at \href{GitHub EPC-3D-Diff}{https://github.com/ALZAHRAALTALIB/EPC-3D-Diff}


\end{abstract}

CT plays an important role in modern radiotherapy by providing accurate anatomical and electron density information required for treatment planning and dose calculation. Nevertheless, during treatment delivery, CBCT is primarily used for patient positioning and image guided radiotherapy \cite{int1}. However, CBCT often suffers from substantial image quality degradation caused by scatter contamination, beam hardening, truncation, detector imperfections, and reconstruction limitations \cite{int2}. These effects lead to inaccurate Hounsfield Units (HU), reduced soft tissue contrast, and spatial distortions \cite{int3}, making CBCT unsuitable for direct dose calculation and many adaptive radiotherapy workflows. Accurate synthetic CT (sCT) generation from CBCT has therefore emerged as a critical problem in adaptive radiotherapy \cite{int3}. The goal is to transform CBCT volumes into CT equivalent images while preserving anatomy and restoring electron density fidelity, enabling dose recalculation and treatment adaptation without additional diagnostic CT acquisitions \cite{int4,int5}, thereby reducing radiation exposure, cost, and workflow complexity. 
In practice, this translation must be robust across acquisition differences, since CBCT is typically cone-beam with distinct scatter and truncation characteristics, while planning CT is commonly acquired under fan-beam/helical geometries with different physics constraints. 

CBCT to sCT synthesis has progressed from handcrafted correction pipelines to learned image translation. Supervised CNN regressors can improve HU fidelity and suppress artifacts but tend to over smooth fine details due to regression to the mean behaviour \cite{Rusanov2022CBCTReview}. GAN based translators were introduced to reduce blurring and better match CT texture statistics, and they have shown strong performance in CBCT to CT conversion for adaptive radiotherapy \cite{Liang2019CycleGAN}. However, adversarial training can be sensitive to optimization choices and may introduce unrealistic high frequency detail, raising concerns about anatomical reliability. 

More recently, conditional diffusion models \cite{altalib2025cond} have emerged as an alternative for CBCT to CT synthesis \cite{altalib2025eqdiff,int12}, providing stable training and iterative refinement of structure and HU values \cite{altalib2026,int18}. Diffusion variants have also included energy guided formulation \cite{int19} and latent space diffusion \cite{Chen2024LDM}. Despite these advances, most CBCT diffusion approaches remain primarily image domain, optimizing voxel space similarity without explicitly enforcing measurement consistency. 
In parallel, equivariant imaging theory shows that leveraging known symmetries can improve identifiability and robustness in inverse problems \cite{ChenTachella2021EI}, but such principles are rarely integrated into CBCT to sCT diffusion pipelines in a way that directly matches CT acquisition physics. Moreover, clinical deployment often requires generalization under limited training data or domain shifts; recent few-shot diffusion approaches address data scarcity but still operate largely in image space \cite{Yeap2025FewShot}. Motivated by these gaps, we propose a novel Equivariant Physics Consistent Conditional 3D Latent Diffusion (\textbf{EPC-3D-Diff}) framework that combines (i) \textbf{Conditional diffusion in a 3D latent space} for scalable volumetric CBCT to CT synthesis, (ii) \textbf{Physics consistent measurement domain constraints} derived from the CT forward model to discourage physically implausible HU solutions, and (iii) \textbf{Explicit rotational equivariance} aligned with operator symmetries to improve robustness under scanner and domain shifts. 

We evaluate EPC-3D-Diff on both phantom and clinical CBCT/CT data, including multiscanner settings and cross domain generalization experiments.

\section{Methodology}

In this section, we present the mathematical models for the CBCT and CT forward operators and the proposed EPC-3D-Diff framework for training sCT image generation from CT volumes using CBCT guidance. 

In the following, we denote $\*x_0, \*x_c\in\R^{N\times M \times S}$ respectively the CT and CBCT aligned 3D volumes in the discretised image domain and $\*A_0\in\R^{(D_0 \cdot \varphi_0) \times (N\cdot M \cdot S)}$, $\*A_c\in\R^{(D_c \cdot \varphi_c) \times (N\cdot M \cdot S)}$ the forward operators representing, respectively, the discretised helical CT and Cone-Beam CT geometries, with $D_{0,c}$ indicating the number of voxels of the 2D detector array and $\varphi_{0, c}$ the angular views that are different according to the geometries. The CT measurements are modeled using the Beer law for mono-energetic X-ray transmission $\*p_{0} = \mathrm{Poisson}\left(I_0e^{-\*A_{0}}\*x_{0}\right)$. After taking the log as in \cite{fessler2000}, we obtain the linear model
\begin{equation}\label{eq:proj}
    \*y_0 = \*A_0\*x_0 + \*n_0, \quad  \*n_0 \sim \+N(0, \*W)
\end{equation}
with $\*n_0$ Gaussian noise with diagonal covariance matrix $\*W$. For CBCT data, we consider a similar model denoted by subscript $c$.

\subsection{Equivariant Physics Consistent Conditional 3D Latent Diffusion} 
Let $\*x_0\in\R^{B\times N\times M\times S}$ denote $B$ high quality CT volume batches and $\*x_c\in\R^{B\times N\times M\times S}$ the corresponding CBCT volumes. Our goal is to learn a conditional generative model that reconstructs a CT quality image from CBCT input while enforcing equivariance consistency with measured projection data, given the mathematical multi operators models of the CT and CBCT data in Eq. (\ref{eq:proj}).

To improve computational efficiency and stabilize training in 3D, we operate in a learned latent space. Let $\+E(\cdot)$ and $\+D(\cdot)$ denote an Autoencoder (AE) and decoder pair. The CT and CBCT volumes are mapped to latent representations:
\begin{equation}
    \*z_0 = \+E_\psi(\*x_0), \, \*z_c = \+E_\psi(\*x_c), \; \+E_\psi:\R^{N \times M \times S}\rightarrow
\R^{N' \times M' \times S},\, N'<N,\, M'<M
\end{equation}
We learn a conditional diffusion model that reconstructs $\*z_0$ from noisy latent observations conditioned on $\*z_c$. We adopt the variance preserving diffusion formulation \cite{ho2020} where a forward Markov process gradually corrupts the latent variable $\*z_c$ over $T$ time steps $q(\*z_t, \*z_0) = \+N(\sqrt{\bar{\alpha}_t}\*z_0, (1 - \bar{\alpha}_t)\*I)$ where $\bar{\alpha}_t := \prod_{s=1}^t (1 - \beta_s)$ and $\{\beta_t\}_{t=1}^T$ is the variance linear schedule. In the forward diffusion process, noisy latent vectors $\*z_t$ at time step $t$ are obtained by adding noise progressively as 
\begin{equation}
    \*z_t=\sqrt{\bar{\alpha}_t} \*z_0 + \sqrt{1 - \bar{\alpha}_t}\boldsymbol{\epsilon}, \quad \boldsymbol{\epsilon}\sim\+N(0,\*I)
\end{equation}

We parameterize the reverse process using a 3D conditional U-Net that predicts the injected noise $\bm\epsilon_{\bm\theta}(\*z_t, t, \+E_\psi(\*x_c))$. 
Given noisy latent $\*z_t$, timestep $t$, and CBCT latent $\*z_c$. It is important to note that EPC-3D-Diff exploits as guidance the CBCT measurements $\*y_c$ that are mapped into the image domain data $\*x_c= \*A_c^T\*y_c$ by applying the transpose of the CBCT physics operator $\*A_c$, which is implemented using the Filtered back projection algorithm. 

From the predicted noise, the 3D latent representation $\hat{\*z}_0$ is obtained as 
\begin{equation}
    \hat{\*z}_0= \frac{\hat{\*z}_t - \sqrt{1 - \bar{\alpha}_t}\bm\epsilon_{\bm\theta}(\*z_t, t, \+E_\psi(\*x_c))}{\sqrt{\bar{\alpha}}_t} 
\end{equation}
At the end of the reverse process, the 3D latent variable $\*z_0$ is mapped to the sCT using the AE decoder $\hat{\*x}_0 = \+D_\omega  (\hat{\*z}_0)$. The training objective related to DDPM is 
\begin{equation}
\mathcal{L}_{\text{DDPM}}(\bm\theta) =
\mathbb{E}_{t, \*x_0, \bm\epsilon}
\|\bm\epsilon - \bm\epsilon_{\bm\theta}(\*z_t, t, \mathcal{E}_\psi(\*x_c))
\|_2^2.
\end{equation}

\paragraph{Projection Equivariant Diffusion Consistency}: Given the CT operator $\*A_0$ under circular acquisition geometry and noiseless condition, for in plane 3D rotations denoted by $\*R_\phi$ on the scanner angular axis $\phi$, the imaging system satisfies the projection based equivariance property which follows from the rotational symmetry of the source–detector trajectory $\*A_0(\*R_\phi \hat{\*x}_0)(\eta) = \*A_0(\*x)(\eta - \phi)$ meaning that the synthetic CT volume generated by rotation $\hat{\*x}_0$, after the forward CBCT operator $\*A_0$, should match the correspondingly translated CT measurements $\*y_0$. Under the Gaussian noise model for $\*y_0$ as specified in Eq. (\ref{eq:proj}), we design the equivariant loss in order to minimize the $L_2$-norm loss by enforcing geometric consistency between the generated sCT $\hat{\*x}_0$ and the measured CT projections $\*y_c$.
\begin{equation}\label{eq:L_equiv}
\mathcal{L}_{\mathrm{eq}} = \sum_{i=1}^{N_{\phi}}\|\*T_{\phi_i}(\*y_0) -  \*A_0\*R_{\phi_i} (\hat{\*x}_0)  \|_2^2,
\end{equation}
with $N_\phi$ the number of angular rotations and $\*T_\phi$ the fractional angular shift. From a theoretical point of view, considering the score function $\nabla_{\*x_t} \log p_{\bm\theta}(\*x_t| \*y_c)$ of the variational DDPM, this corresponds to an augmented posterior
\begin{equation}\label{eq:post_equiv}
p_{\bm\theta}(\*x_0 | \*x_t, \*y_0, \*y_c) \propto p_{\bm\theta}(\*x_t | \*y_c) \exp\Big(-\lambda_\text{eq} \| \*T_\phi(\*y_0) - \*A_0\*R_\phi (\hat{\*x}_0)  \|_2^2 \Big),
\end{equation}
with score function 
\begin{equation}\label{eq:score_nabla}
\nabla_{\*x_t} \log p_{\bm\theta}(\*x_0 | \*x_t, \*y_0) \approx \nabla_{x_t} \log p_{\bm\theta}(\*x_t | \*y_c) - \frac{\lambda_\text{eq}}{\sigma^2} \*R_\phi^T\*A_0^T(\*T_\phi(\*y_0) - \*A_0\*R_\phi (\hat{\*x}_0))
\end{equation}
assuming Gaussian measurements noise with variance $\sigma^2$. 

From a regularization perspective, the posterior (\ref{eq:post_equiv}) introduces a physics-informed prior that penalizes violations of acquisition symmetry. This can be interpreted as projecting intermediate diffusion samples onto an equivariant sub manifold defined by the $\*A_0$ operator symmetry. 

To improve structural fidelity, we add complementary image domain losses:
\begin{equation}
\mathcal{L}_{\text{L}_1} = \|\hat{\*x}_0-\*x_0\|_1, \quad
\mathcal{L}_{\text{edge}} = \|\nabla \hat{\*x}_0 - \nabla \*x_0\|_1, \quad
\mathcal{L}_{\text{lap}} = \|\Delta \hat{\*x}_0 - \Delta \*x_0\|_1
\end{equation}
where $L_1$ preserves voxel level accuracy, the gradient loss $\mathcal{L}_\text{edge}$ preserves structural edges with $\nabla$ denoting a 3D finite difference gradient operator, and the Laplacian loss $\mathcal{L}_\text{lap}$ preserves high frequency details and fine anatomical structures with $\Delta$ spatial displacement. The total training loss is a weighted sum: 
\begin{equation}\label{eq:EPC_loss}
\mathcal{L}(\bm\theta)
= \mathcal{L}_{\text{DDPM}}(\bm\theta)
+
\lambda_1 \mathcal{L}_{\text{L}_1}
+
\lambda_2 \mathcal{L}_{\text{edge}}
+
\lambda_3 \mathcal{L}_{\text{lap}}
+
\lambda_{\text{eq}}\mathcal{L}_{\text{eq}}.
\end{equation}

\begin{figure}[!h]
    \centering
    \includegraphics[width=\textwidth]{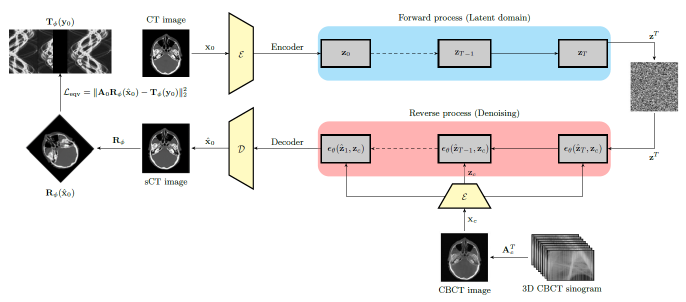}
	\caption{EPC-3D-Diff with projection equivariant loss. The sCT $\hat{\*x}_0$ is rotated via $\*R_\phi$ to enforce rotational consistency. The CT forward operator $\*A_0$ and rotation operator $T_\phi$ are applied in the projection domain for the equivariant loss $\mathcal{L}_{\mathrm{eq}}$.}\label{fig:EPC_workflow}
\end{figure}

\noindent The diagram in Fig. \ref{fig:EPC_workflow} shows the workflow of the forward and reverse diffusion process in the latent domain with conditional CBCT guidance for EPC-3D-Diff. 

\paragraph{Latent Autoencoder Pretraining and 3D Conditional U-Net}\label{sec:lat_enc}: The encoder $\+E_{\bm\psi}(\cdot)$ and decoder $\+D_{\bm\omega}(\cdot)$ each consist of 3D residual blocks with group normalization and SiLU activation, with downsampling in the encoder and upsampling in the decoder. 
Skip connections within each block stabilize gradient flow. The network is lightweight, memory efficient, and preserves spatial fidelity. The AE pretraining loss combines a $L_1$ reconstruction term with an edge preserving regularizer: 
\begin{equation}
\mathcal{L}_{\mathrm{AE}} = \| \+D_{\bm\omega}(\+E_{\bm\psi}(\*x_0)) - \bar{\*x}_0 \|_1 + \lambda \, \|\nabla \+D_{\bm\omega}(\+E_{\bm\psi}(\*x_0)) - \nabla \bar{\*x}_0\|_1,
\end{equation}
where $\*x_0$ is the ground truth 3D CT volume, $\bar{\*x}_0$ is the reconstructed volume, and $\lambda$ balances reconstruction fidelity and anatomical edge sharpness. After convergence of the AE training, the encoder and decoder parameters $\bm\psi$ and $\bm\omega$ are fixed. The diffusion model is then trained exclusively in the latent space using the pre-trained encoder with reduced computational complexity. 
 
The noise prediction network $\bm\epsilon_{\bm\theta}$ is a 3D conditional U-Net with residual blocks $h_0 = \mathrm{Conv}_{3\times 3}([\*z_t \Vert \*z_c])$, $h_i = \mathrm{ResBlock}_i(h_{i-1}, \mathbf{t})$, where $[\cdot \Vert \cdot]$ denotes channel wise concatenation, $i=1,\dots,L$, group normalization, SiLU activation, down-sampling in the encoder and symmetric up-sampling in the decoder $\hat{\epsilon}_\theta = \mathrm{Conv}_{3\times 3}(h_L)$ and $\mathbf{t} = \mathrm{MLP}(\gamma(t))$ is the sinusoidal timestep embedding. Skip connections are used within each residual block to stabilize gradients.

\paragraph{EPC-3D-Diff Inference Strategy}: At test time, only the CBCT input $\*x_c$ is available. Therefore, sampling proceeds from the learned conditional prior $\*x \sim p_{\bm\theta}(\*x \mid \*x_c)$ using deterministic DDIM updates $\hat{\*z}_{t-1}=\sqrt{\bar{\alpha}_{t-1}} \left( \frac{\hat{\*z}_t - \sqrt{1 - \bar{\alpha}_t}\boldsymbol{\epsilon}_t}{\sqrt{\bar{\alpha}}_t} \right) + \sqrt{1 - \bar{\alpha}_{t-1}}\boldsymbol{\epsilon}_t$. Unlike classical CT data-consistency, $\hat{\*x}_0$ and $\*y_c$ are cross modality, enabling learning a physics informed conditional diffusion prior at training, while inference remains operator free and computationally efficient.

\section{Experiments and Results}

\subsection{Datasets, Preprocessing and Implementation Details}

We evaluated EPC-3D-Diff on two diverse head and neck data. The NWH phantom dataset from the Ninewells (Dundee) hospital contains 10 patients CBCT/CT raw and image domain data, split patient wise into 8 for training and 2 for testing. Each patient provides 92 paired CBCT/CT slices, yielding 736 training pairs and 184 testing pairs. The JUST clinical study with ethical approval Mar2026/191-38 at King Abdullah University Hospital (Jordan) contains 14 patients with high diversity in terms of demographics and gender, split patient wise into 11 for training (631 slices) and 3 for testing (173 slices). CBCT and CT volumes were paired slice wise after correcting for possible stack reversal and in plane offsets using correlation based translation estimation. All images were converted to HU, clipped to fixed window $[-1000, 2000]$ HU, and linearly normalized to $[-1,1]$. To ensure consistent input size, paired slices were jointly cropped using a foreground mask and resized to $256\times256$. 

All models were implemented in PyTorch using 2 NVIDIA A5000 GPUs; the latent AE dimensionality was set to $C=4$. The EPC-3D-Diff training was performed with AE fixed pretrained weights $\bm\omega$ and $\bm\psi$ as described in Section \ref{sec:lat_enc}. The conditional 3D U-Net was operating in latent space with base width 64 and timestep embedding dimension 256. Training used a forward diffusion process with $T=1000$ time steps and a linear noise schedule $\beta_t \in [10^{-4}, 5\times10^{-3}]$. Batch size of 2 was used, and when training jointly on NWH and JUST we employed balanced mini batches to prevent dominance of the larger cohort. Optimization was performed with Adam using a learning rate of $10^{-5}$ and training with 2500 epochs. The weights of the loss in Eq. (\ref{eq:EPC_loss}) were set as $\lambda_1=0.6$, $\lambda_2=0.2$, $\lambda_3= 0.2$ and $\lambda_\texttt{eq}=0.1$. Projection domain equivariance regularization was enabled and applied every 10 epochs using two random rotations per sample with adaptive error norm $\lambda_\texttt{eq}$. For inference, we used a DDIM  with 100 steps to accelerate sampling with the same trained network.

\begin{figure}[!h]
	\includegraphics[width=\textwidth]{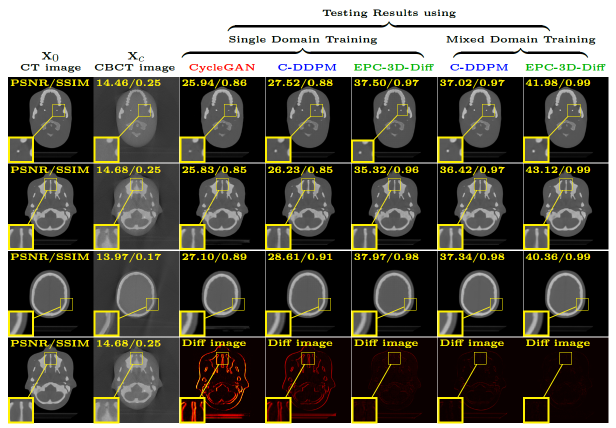}
	\caption{Qualitative testing comparison on NWH phantom test set. Columns show the reference CT ($\*x_0$), CBCT ($\*x_c$), and sCT results for single domain (NWH) and mixed domain (NWH+JUST) training. Overlaid values report PSNR/SSIM relative to $\*x_0$. The bottom row shows absolute difference maps $|\*x_0-\hat{\*x}_0|$.}\label{fig:NWH}
\end{figure}

\subsection{Comparison with different Datasets and Scanners} 

All methods were quantitatively evaluated in terms of mean $\pm$ std over the test pairs using SSIM, PSNR between sCT and reference CT, along with the MSE, MAE, and HU profiles. To assess robustness to domain shifts such as scanner geometries,  we compare CycleGAN~\cite{Liang2019CycleGAN}, a baseline conditional diffusion model (C-DDPM without equivariance), and the proposed EPC-3D-Diff under two training regimes: (i) \textbf{single domain training} (train and test on the same dataset) and (ii) \textbf{mixed domain training} (train on NWH+JUST and test on each dataset separately). Fig. \ref{fig:NWH} presents representative NWH test slices for all methods in both training regimes. Visually, mixed domain training yields sharper anatomical boundaries and fewer HU dependent distortions, particularly in challenging regions with heterogeneous intensities and fine structures. 

This trend is consistent with the quantitative summary in Table \ref{table:NWH} (mean $\pm$ std in the testing cohort), where EPC-3D-Diff achieves the highest fidelity to the CT reference. In particular, multi domain training improves the performance of NWH test compared with training on NWH alone (PSNR: $38.44$ vs.\ $30.99$~dB; SSIM: $0.99$ vs.\ $0.92$), indicating that additional training diversity improves generalization even when cohorts are acquired under different scanner specific setups. In addition, EPC-3D-Diff improves noticeably in the multiple domain ($\sim + 7$ dB) compared to C-DDPM because it encourages invariances that are beneficial in domain shifts.  Because the two cohorts are not perfectly balanced, we use balanced mini batches to prevent the larger cohort from dominating optimization. The consistent gains in Fig. \ref{fig:NWH} and Table \ref{table:NWH} suggest that the improvement is driven by the diversity of the domains rather than the size of the dataset.

\begin{table}[!h]
\centering
\caption{Quantitative results (mean $\pm$ std) on the NWH phantom test set on single and multiple domain training for CycleGAN, C-DDPM, and EPC-3D-Diff.}\label{table:NWH}
\resizebox{\textwidth}{!}{
\begin{tabular}{lp{2cm}>{\centering\arraybackslash}p{2.2cm}>{\centering\arraybackslash}p{2.2cm}>{\centering\arraybackslash}p{2.2cm}>{\centering\arraybackslash}p{2.2cm}}
\toprule
& \bf Method  
& \bf PSNR (dB)
& \bf SSIM 
& \bf MAE (dB)
& \bf MSE (dB)\\
\midrule
\midrule
& CBCT - CT & 14.41 $\pm$ 0.351 & 0.26 $\pm$ 0.07  & -22.30 $\pm$ 2.80  & -33.14 $\pm$ 0.37  \\ 
\midrule
\multirow{3}{1.5cm}{Single Domain Training} & \textcolor{red}{CycleGAN}  & 25.85 $\pm$ 7.82 & 0.82 $\pm$ 0.09 & -40.67 $\pm$ 2.83 & -62.38 $\pm$ 4.78 \\
\cmidrule(lr){2-6}
& \textcolor{blue}{C-DDPM} & 30.77 $\pm$ 4.31 & 0.91 $\pm$ 0.05  & -21.27 $\pm$ 0.2\; & -66.53 $\pm$ 3.53  \\
&  \textcolor{green!70!black}{EPC-3D-Diff} &  \textbf{30.99} $\pm$ \textbf{4.21}  & \textbf {0.92} $\pm$ \textbf{0.05}  & \textbf{-21.34} $\pm$ \textbf{0.2}\;\, & \textbf{-66.69} $\pm$ \textbf{5.38} \\
\midrule
\midrule
\multirow{3}{1.5cm}{Multiple Domain Training} & \textcolor{red}{CycleGAN} & 30.50 $\pm$ 7.27 & 0.88 $\pm$ 0.07 & -42.67 $\pm$ 2.64 & -62.35 $\pm$ 4.65 \\
\cmidrule(lr){2-6}
& \textcolor{blue}{C-DDPM}  & {31.00} $\pm$ 4.48 & 0.93$\pm$ 0.05 & -44.52 $\pm$ 2.0\;\,  & -66.69 $\pm$ 3.66 \\
&  \textcolor{green!70!black}{EPC-3D-Diff} & \textbf{38.44} $\pm$ \textbf{2.28} & \textbf{0.99} $\pm$ \textbf{0.01}  & \textbf{-51.70} $\pm$ \textbf{1.05} & \textbf{-86.86} $\pm$ \textbf{3.39} \\
\bottomrule
\end{tabular} 
}
\end{table}

\begin{figure}[!h]
	\includegraphics[width=\textwidth]{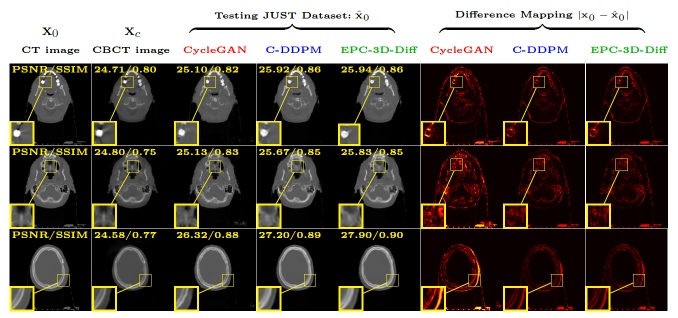}
	\caption{Qualitative testing comparison on JUST clinical test set. Columns show the reference CT ($\*x_0$), CBCT ($\*x_c$), sCT ($\hat{\*x}_0$) and corresponding difference maps.}\label{fig:JUST}
\end{figure}

\begin{figure}[!h]
\centering
\includegraphics[width=\textwidth]{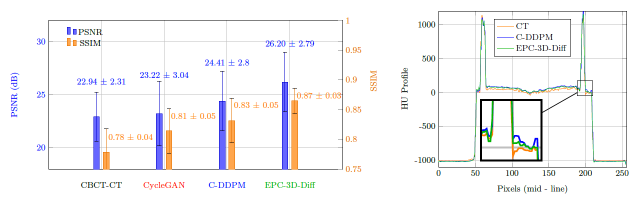}
\caption{Quantitative analysis on JUST clinical test set. \textbf{Left:} Hounsfield unit (HU) line profile extracted along the mid line of a representative slice. \textbf{Right:} Aggregate performance obtained by CycleGAN, C-DDPM, and EPC-3D-Diff.}\label{fig:HU}
\end{figure}

In Fig. \ref{fig:JUST} we include the testing results for JUST clinical single domain training set, where we consistently show how EPC-3D-Diff improves $\sim 1.8$ dB PSNR on average over C-DDPM and across different regions. This is confirmed in Fig. \ref{fig:HU} (left) by the quantitative improvement for PSNR and SSIM and lower variability and (right) by the HU profile of the middle line of the sample image. 

In terms of computational time, EPC-3D-Diff requires one forward projection per training epoch with complexity $\mathcal{O}(N M S)$. Since the $\+L_\texttt{eq}$ loss is applied every 10 epochs, the overhead is less than 15\% of the total training time. Inference, no projection operations are required, resulting in 58 seconds runtime per patient.

\section{Conclusion}

We proposed EPC-3D-Diff, a novel projection equivariant 3D conditional latent diffusion framework for CBCT to sCT synthesis that embeds acquisition physics as a structural inductive bias. By enforcing projection domain rotational consistency during training, the model improves HU fidelity while enhancing robustness to mixed domain training and scanner geometry shifts. Results on phantom and clinical head and neck data show proposed approach consistently outperforms conventional conditional diffusion baselines. Importantly, physics operators are required only during training enabling fast and practical deployment.

\section*{Acknowledgment}

A. Altalib is supported by the Jordan University of Science and Technology PhD scholarship and A. Perelli by the Royal Academy of Engineering / Leverhulme Trust Research Fellowship LTRF-2324-20-160. 

The authors thank Sandra and Yamen Alhyari for their contribution to improve the manuscript and Christopher Taylor and Sankar Pillai for supporting in the phantom data collection at Ninewells hospital.

\bibliographystyle{splncs04}
\bibliography{ref}

\end{document}